\documentclass[11pt]{article}

\usepackage[preprint]{acl}

\usepackage{times}
\usepackage{latexsym}

\usepackage[T1]{fontenc}

\usepackage[utf8]{inputenc}

\usepackage{microtype}

\usepackage{inconsolata}

\usepackage{graphicx}
\usepackage{adjustbox}
\usepackage{amssymb}
\usepackage{multirow}
\usepackage[table]{xcolor}
\usepackage{algorithm}
\usepackage{algpseudocode}
\usepackage{amsmath}
\usepackage{makecell}
\usepackage{enumitem} 

\usepackage{xcolor}

\title{EvolMem: A Cognitive-Driven Benchmark for Multi-Session Dialogue Memory}

\author{
    Ye Shen$^{1,2}$,
    Dun Pei$^{1,2}$,
    Yiqiu Guo$^{2,3}$,
    Junying Wang$^{2,3}$,
    Yijin Guo$^{1,2}$,\\
    \textbf{Zicheng Zhang}$^{2}$,
    \textbf{Qi Jia}$^{2,}$\thanks{Corresponding authors.},
    \textbf{Jun Zhou}$^{1}$, 
    \textbf{Guangtao Zhai}$^{1,2,}$\footnotemark[1]\\
    \textsuperscript{\rm 1}Shanghai Jiao Tong University,
    \textsuperscript{\rm 2}Shanghai Artificial Intelligence Laboratory,\\
    \textsuperscript{\rm 3}Fudan University\\
    \texttt{\{sherylye,dpei29\}@sjtu.edu.cn, jiaqi@pjlab.org.cn}
}

\begin{document}
\maketitle
\begin{abstract}

Despite recent advances in understanding and leveraging long-range conversational memory, existing benchmarks still lack systematic evaluation of large language models(LLMs) across diverse memory dimensions, particularly in multi-session settings. In this work, we propose EvolMem, a new benchmark for assessing multi-session memory capabilities of LLMs and agent systems. EvolMem is grounded in cognitive psychology and encompasses both declarative and non-declarative memory, further decomposed into multiple fine-grained abilities. To construct the benchmark, we introduce a hybrid data synthesis framework that consists of topic-initiated generation and narrative-inspired transformations. This framework enables scalable generation of multi-session conversations with controllable complexity, accompanied by sample-specific evaluation guidelines. Extensive evaluation reveals that no LLM consistently outperforms others across all memory dimensions. Moreover, agent memory mechanisms do not necessarily enhance LLMs’ capabilities and often exhibit notable efficiency limitations. Data and code will be released at \url{https://github.com/shenye7436/EvolMem}.

\end{abstract}

\section{Introduction}

As large language models (LLMs) are increasingly deployed in interactive and long-horizon applications~\citep{agent,memgpt,longtext,interaction}, memory plays a central role in enabling models to preserve user preferences, accumulate knowledge, and support coherent long-term interactions. Consequently, rigorous evaluations of memory capabilities, particularly the ability to maintain and utilize information across sessions, are indispensable for diagnosing model limitations and guiding future improvements.

Although there is a substantial body of research on memory in LLMs and agent systems~\citep{llmagentsurvey}, important limitations remain in existing evaluation paradigms. First, most existing evaluations assess memory either within a single session~\citep{multichallenge,memoryagentbench} or through single-turn long-context inputs~\cite{ruler,nolima,perltqa}. Such settings fall short of realistic usage, where models are expected to accumulate, retain, and utilize information across multiple conversations with users. Multi-session dialogue more closely reflects real-world interactions and poses substantially greater challenges for long-term memory understanding and management. Second, current evaluations place a strong emphasis on recalling previously provided information, which primarily corresponds to declarative memory. From a cognitive psychology perspective, however, human memory also includes non-declarative systems that rely on distinct mechanisms and support habits and implicit knowledge~\citep{twomemory}. Existing dialogue benchmarks~\citep{dialsim,longmemeval,personamem,multichallenge} largely overlook non-declarative memory and lack a comprehensive set of assessment dimensions, leaving important aspects of models’ memory capabilities insufficiently evaluated.

\begin{figure}[t]
\includegraphics[width=\linewidth]{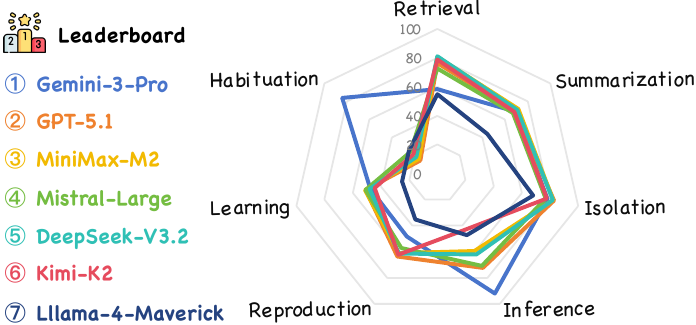}
  \caption{The leaderboard and capability distribution of LLMs across fine-grained memory dimensions.}
  \label{fig:llm_leaderboard}
\end{figure}

\newcommand{\coloredblock}[1]{\textcolor{#1}{\rule{3mm}{3mm}}}

\definecolor{RT}{HTML}{F27970}
\definecolor{SM}{HTML}{BB9727}
\definecolor{IS}{HTML}{54B345}
\definecolor{IF}{HTML}{32B897}
\definecolor{RP}{HTML}{05B9E2}
\definecolor{HB}{HTML}{8983BF}
\definecolor{PL}{HTML}{ECA8A9}

\begin{table*}[t]
\centering
\begin{adjustbox}{max width=\textwidth}
\setlength{\tabcolsep}{2pt}

\begin{tabular}{l|c|c|c|c c c c c c c|c} 
\hline

\textbf{Benchmark} & \textbf{Multi-turn}&\textbf{ Multi-session} & \textbf{Query Type} & \multicolumn{7}{c|}{\textbf{Memory Types}}  &\textbf{Diverse Metrics} \\

\hline

NIAH~\citep{niah}   & $\times$ & $\times$ & Obj. & $\coloredblock{RT}$ &  &  &  &  & & & - \\

Ruler~\citep{ruler} & $\times$ & $\times$ & Obj. & $\coloredblock{RT}$ & $\coloredblock{SM}$ & $\coloredblock{IS}$ & $\coloredblock{IF}$ &  & & & $\checkmark$ \\

NoliMa~\citep{nolima}  & $\times$ & $\times$ & Obj. & $\coloredblock{RT}$ & $\coloredblock{IF}$ &  &  &  & &  & $\times$\\

Multi-Session Chat~\citep{msc}  & $\checkmark$ & $\checkmark$ & Subj. & $\coloredblock{RT}$& $\coloredblock{SM}$ & & &  &  &  & $\checkmark$ \\

PerLTQA~\citep{perltqa}  & $\checkmark$ & $\times$ & Obj. & $\coloredblock{RT}$ & $\coloredblock{SM}$ & $\coloredblock{IS}$ &  & & & & $\checkmark$ \\

LongMemEval~\citep{longmemeval}   & $\checkmark$ & $\checkmark$ & Obj. & $\coloredblock{RT}$ & $\coloredblock{IF}$ &  &  &  & & & $\times$\\

PersonaMem~\citep{personamem}  & $\checkmark$ & $\checkmark$ & Subj. & $\coloredblock{RT}$ & $\coloredblock{SM}$ & $\coloredblock{IF}$ &  & & &   & $\times$\\
        
Multichallenge~\citep{multichallenge}  & $\checkmark$ & $\times$ & Obj. \& Subj. & $\coloredblock{RT}$ & $\coloredblock{SM}$ & $\coloredblock{IF}$ && & &  & $\checkmark$\\

MemoryAgentBench~\citep{memoryagentbench}   & $\checkmark$ & $\times$ & Obj. \& Subj. & $\coloredblock{RT}$ & $\coloredblock{SM}$ & $\coloredblock{IF}$ & & &  & & $\checkmark$ \\
\hline
\textbf{EvolMem (Ours)} & $\checkmark$ &$\checkmark$ & Obj. \& Subj. & $\coloredblock{RT}$ & $\coloredblock{SM}$ & $\coloredblock{IS}$ & $\coloredblock{IF}$ & $\coloredblock{RP}$ & $\coloredblock{PL}$ & $\coloredblock{HB}$  & $\checkmark$\\
\hline
\end{tabular}
\end{adjustbox}

\caption{A comparison of memory evaluation benchmarks. \textbf{Multi-turn} represents whether the benchmark follows a multi-turn format. \textbf{Multi-session} represents whether it follows a multi-session format. \textbf{Query type} represents the types of questions to assess model memory ability, and Obj. means Objective while Subj. means Subjective. Type of Memory represents the memory capabilities involved in the benchmark. \coloredblock{RT} represents retrieval, \coloredblock{SM} represents summarization, \coloredblock{IS} represents isolation, \coloredblock{IF} represents inference, \coloredblock{RP} represents reproduction, \coloredblock{PL} represents learning and \coloredblock{HB} represents habituation. \textbf{Diverse Metrics} indicates the use of diverse evaluation metrics. }
\label{tab:your_label}
\end{table*}

In this work, to bridge these gaps, we propose a memory benchmark, EvolMem, for assessing LLMs and agent systems. EvolMem focuses on multi-session interactions to capture the evolving development of long-term memory through successive interactions. Grounded in cognitive science, it offers a comprehensive evaluation framework that covers both declarative and non-declarative memory, and further operationalizes seven fine-grained abilities including retrieval, inference and etc.

We introduce a hybrid data synthesis framework that integrates topic-initiated generation and narrative-inspired transformation. Topic-initiated generation produces structured overviews from a wide range of topics, while narrative-inspired transformation converts abundant long-context benchmarks into multi-session dialogues via segmentation, ensuring interaction coherence alongside linguistic and topical diversity. Subsequently, we apply filtering and challenge injection to create scenarios that systematically induce complex memory demands. The resulting dialogues support customized evaluation of different memory abilities, including result-oriented, process-oriented, and open-ended metrics.

Through a systematic evaluation of state-of-the-art LLMs and agent systems, we effectively differentiate their long-term memory capabilities and pinpoint specific vulnerabilities in memory tasks. Notably, neither LLMs nor agent memory mechanisms achieve consistent leading performance among fine-grained memory capabilities. We also observe systematic weaknesses in non-declarative memory, even among advanced models, highlighting limitations not captured by existing benchmarks. Moreover, agent memory mechanisms introduce efficiency bottlenecks, particularly  when encountering long conversation histories.

In summary, our contributions are:
\begin{itemize}
    \item We propose a cognitive-driven memory benchmark, enabling comprehensive evaluation of LLMs' memory capabilities.
    \item We develop a multi-source data synthesis framework to enhance the benchmark’s diversity, and introduce multi-faceted evaluation metrics for systematic evaluation.
    \item Our evaluation indicates that comprehensive superiority in memory capabilities remains challenging for both LLMs and agents, pointing out multiple future directions.

\end{itemize}

\section{Related Work}

\subsection{The Evolution of LLMs \& Agents}
 Conventional LLMs are constrained by fixed context windows, limiting long-horizon reasoning and multi-session consistency~\citep{liu2024lost}.With the expansion of context capabilities, state-of-the-art LLMs such as GPT-5~\citep{gpt5}, Gemini 3.0 Pro~\citep{gemini3pro}, and DeepSeek-V3.2~\citep{deepseek} demonstrate enhanced capabilities in these memory-intensive tasks.

\begin{figure*}[t]
\includegraphics[width=\textwidth]{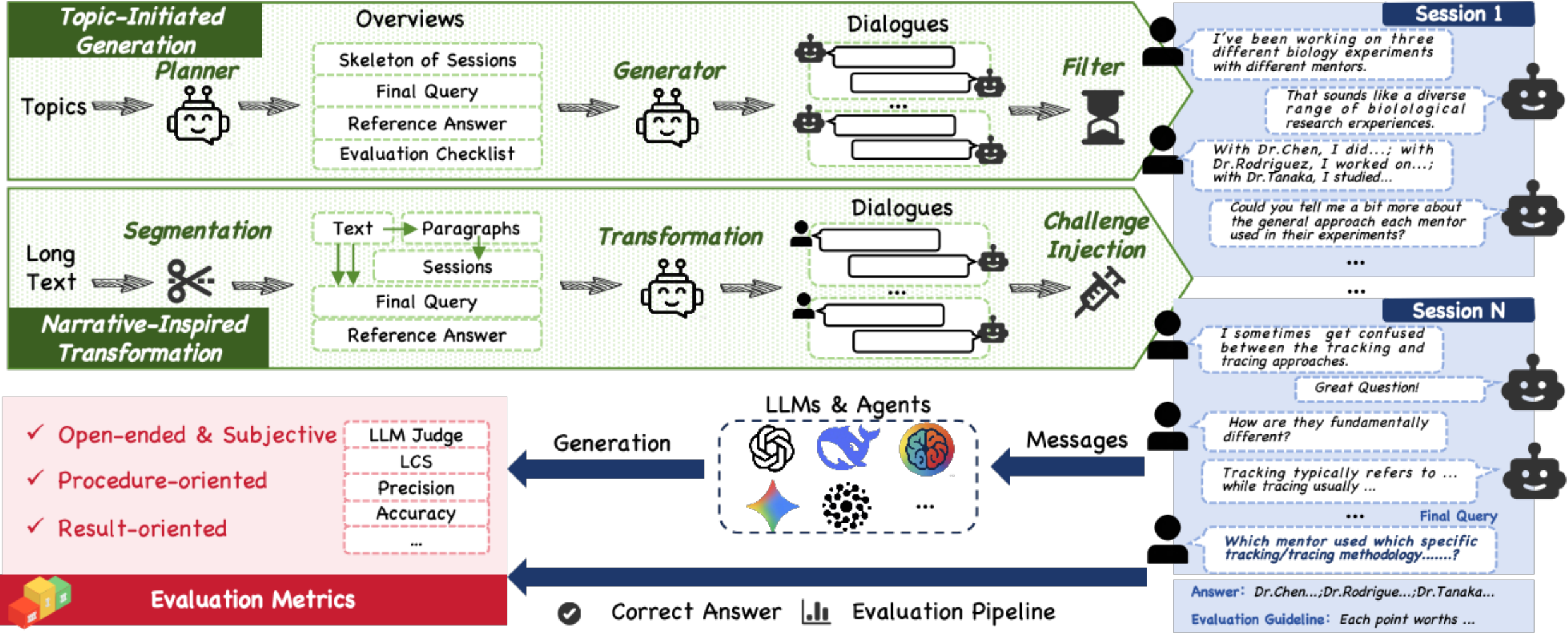}
  \caption {The outline of our hybrid data synthesis framework and adopted evaluation metrics.}
  \label{fig:approach}
\end{figure*}

As models evolve into general-purpose autonomous agents, memory mechanisms are pivotal for stable reasoning and persistent execution. Existing approaches are broadly categorized into structure-augmented RAG (e.g., HippoRAG~\citep{hipporag}, Mem0~\citep{mem0}), which augments retrieval via structured representations, and agentic memory (e.g., A-Mem~\citep{amem}, MemoryOS~\citep{memoryos}), which orchestrates memory through autonomously decided retrieval-reflection-update loops. Given memory's critical role in agentic evolution, rigorous benchmarking is imperative.

\subsection{Memory Evaluation Benchmarks}
Long-text benchmarks such as NIAH~\citep{niah}, Ruler~\citep{ruler}, and Nolima~\citep{nolima} test the ability to retain single-turn information. Limited to fixed context windows, these benchmarks fail to reflect the evolution of information. Furthermore, PerLTQA~\citep{perltqa}, Multichallenge~\citep{multichallenge}, and MemoryAgentBench~\citep{memoryagentbench} employ a multi-turn format closer to the basic pattern of interaction than long text input. However, information within one session still primarily tests short-term memory.

 Subsequently, benchmarks like DiaSim~\citep{dialsim}, LongMemEval~\citep{longmemeval}, PersonaMem~\citep{personamem}, Multi-Session Chat~\citep{msc}, and Conversation Chronicles~\citep{conversation_chronicles} extend to multiple independent but related sessions, mirroring real-world scenarios, and assessing the memory beyond the context window. Nevertheless, they neglect non-declarative memory, which is equally important to examine long-term memory and information retention mechanisms~\citep{human}. To this end, there is an urgent need for benchmarks featuring multi-session dialogues that enable comprehensive evaluation across different memory types.

\section{EvolMem}

An overview of the construction of EvolMem is illustrated in Fig.~\ref{fig:approach}. We begin by classifying memory into declarative and non-declarative types and further subdivide them into fine-grained capabilities. To generate multi-session dialogues, we develop a hybrid synthesis framework that combines topic-initiated generation and narrative-inspired transformation, ensuring diverse topics, coherent multi-session interactions, and rich linguistic content. Finally, we design customized evaluation metrics to assess model performance across different memory abilities, including result-oriented, process-oriented, and open-ended metrics.

\subsection{Memory Classification}
\label{sec:memory_classification}

According to human cognitive psychology, memory is categorized into declarative memory and non-declarative memory~\citep{twomemory}. We explain the definitions and delineate the fine-grained capabilities for each type as follows.

\textbf{Declarative memory} captures the ability to explicitly store and retrieve facts, events, and states. It encompasses knowledge and experiences that can be consciously recalled and manipulated, and is further subdivided into the following capabilities: 

\begin{itemize}[leftmargin=0.5cm, itemsep=0pt, topsep=0pt]
    \item \textbf{Retrieval}~\citep{contentretrieval} involves grasping previously acquired key information amid substantial irrelevant content.
    
    \item \textbf{Summarization}~\citep{planrevision} updates memory representations as goals or task conditions change.
    
    \item \textbf{Isolation}~\citep{informationisolation} requires distinguishing the sources of stored information to prevent interference or source-misattribution errors.
    
    \item \textbf{Inference}~\citep{perceptioninference} requires reasoning about potential relationships and conclusions from saliently contradictory cues hidden in historical messages.
    
    \item \textbf{Reproduction}~\citep{step-by-step-inversion} reconstructs a process demonstrated in historical interactions.
\end{itemize}

\textbf{Non-declarative memory}, in contrast, captures implicit and automated learning that shapes behavior and decision-making without conscious recall. We focus on two core capabilities:

\begin{itemize}[leftmargin=0.5cm,itemsep=0pt, topsep=0pt]
\item \textbf{Learning}~\citep{procedurallearning} refers to internalizing operational rules through repeated experience and executing these rules reliably in new situations.
\item \textbf{Habituation}~\citep{habituation} involves the gradual formation of stable, automatic memory patterns that maintain earlier user instructions across interactions.
\end{itemize}

\subsection{Dialogue synthesis framework}
Topic-Initiated Generation and Narrative-Inspired Transformation leverage rich seed source to enhance data diversity, regulate inter-session dialogue progression through coarse-to-fine paradigms, and modulate difficulty via filtering or injection mechanisms. More details corresponding to each capability are provided in the Appendix~\ref{sec:capability}.

\subsubsection{Topic-Initiated Generation}

\textbf{Seed Data} as initial input comprises $N$ predefined task categories $C_{i}$, each targeting specific capabilities and corresponding requirements, alongside a diverse set of topics $t_{i}$ provided in the Appendix~\ref{sec:topics}. It serves as the foundation for detailed dialogue generation, formally represented as:
\begin{equation}
\mathcal{S}_{\rm TIG} = \{s_i\}_{i=1}^{N}=\left\{ \left( C_i, t_i \right) \right\}_{i=1}^{N},
\end{equation}
where each task category $C_i$ is paired with a systematically grouped set of topics $t_i$ to ensure the sufficient diverse and difficulty in dialogue content.

\textbf{Planner} to synthesize dialogue overviews $\mathcal{O}$ employs a generative function $f_{\rm LLM}({s}_{i}, \mathcal{R}_{\rm PLA})$, where $\mathcal{R}_{\rm PLA}$ represents a set of structural constraints. To ensure sufficient coverage across diverse task types, the planner invokes model to produce $K$ samples per topic, constructing multi-session dialogues with at least $J \geq j$ sessions. Each session is required to include a brief summary $M$ grounded in seed data and specific difficulty, along with a specified number of turns $T$. The model is further instructed to draft a final query $Q$, its corresponding reference answer $A$, and an evaluation checklist $E$ derived from the overview.

\algnewcommand{\Input}{\item[\textbf{Input:}]}
\algnewcommand{\Output}{\item[\textbf{Output:}]}

\begin{algorithm}
\caption{Planner: Overview Generation}
\label{alg:planner}
\begin{algorithmic}[1]

\Input $\mathcal{S}_{\rm TIG}$, $f_{\rm LLM}(\cdot)$
\Output Generated Dialogue Overviews $\mathcal{O}$

\State $\mathcal{O} \gets \emptyset$
\For{$i = 1$ \textbf{to} $N$} \Comment{Iterate over tasks}
    \For{$k = 1$ \textbf{to} $K$} \Comment{Invoke model}
        \State $\mathcal{R}_{\rm PLA} \gets \{ J \ge j, T, M, Q, A, E \}$
       \State $O_{i}^k \gets f_{\text{LLM}}({s}_{i}, \mathcal{R}_{\rm PLA})$ \Comment{Generate Overview}
        \State $\mathcal{O} \gets \mathcal{O} \cup \{O_{i}^k\}$
    \EndFor
\EndFor
\State \Return $\mathcal{O}$
\end{algorithmic}
\end{algorithm}

\textbf{Generator} employs multiple $f_{\rm LLM} (O_{i}^k, \mathcal{R}_{\rm GEN})$ to generate multi-session dialogues $\mathcal{D}$ and complete $Q$, $A$, and $E$, where $\mathcal{R}_{\rm GEN}$ represents natural expansion constraints. By leveraging the heterogeneous linguistic styles, reasoning behaviors, and memory processing capabilities of different models, this approach enables the generation of more diverse and challenging dialogue scenarios. Consequently, the resulting corpus provides a robust and reliable foundation for subsequent evaluation.

\textbf{Filtering} ensures the complexity of our benchmark. Specifically, we use DeepSeek-V3.2~\citep{deepseek} to generate answers, run the evaluation pipeline, and remove all dialogues with scores greater than or equal to 0.8, thus retaining more challenging and discriminative samples.

\subsubsection{Narrative-Inspired Transformation}

\textbf{Seed Data}\quad Long text data is enormous and evaluation methods are nearing saturation~\citep{longbench}, while multi-session dialogues are relatively scarce~\citep{scarce} and more closely reflect real scenarios due to cross-turn interaction logic and contextual coherence. Therefore, we propose a strategy to transform $W$ long text ($L$) from existing benchmark into multi-session dialogues.

\textbf{Segmentation} proceeds in two stages. Firstly, the original long text is partitioned into fine-grained semantic paragraphs $P$, subject to predefined session ($J$) and turn ($T$) constraints. Secondly, $T$ paragraphs are systematically assigned to independent session units. By precisely modulating the distribution of sessions and turns, it effectively simulates real challenges like context truncation, memory decay, and sparse information dispersion. 
\begin{equation}
\mathcal{S}_{\rm NIT} =\{s_{l}\}_{l=1}^{W}= \{ (L_l, P_l )\}_{l=1}^{W}\end{equation}
where each $L_i$ is paired with $P_i$, and $\text{Seed}_{\text{NIT}}$ represents the set of user statements. $Q$\&$A$ can be obtained from the source while the evaluation pipeline ($E$) is established through direct comparison.

\textbf{Transformation}  encapsulates the segmented paragraphs and natural responses into a dialogue format. Specifically, we employ $f_{\text{LLM}}(s_{l}, \mathcal{R}_{\rm NIT})$ to generate human-like agent responses, where $\mathcal{R}_{\rm NIT}$ denotes the constraints for naturalness and coherence. Finally, multi-session dialogues comprising $J$ sessions and $T$ turns are integrated into $\mathcal{D}$, effectively preserving the semantic integrity of the original source.

\textbf{Challenge Injection} employs diverse strategies to stress-test model robustness, including (1) the strategic insertion of contextually plausible yet logically antithetical information to probe the model's consistency of memory, and (2) the integration of established, high-difficulty benchmarks as rigorous evaluation anchors. Overall, this configuration necessitates non-linear reasoning, requiring the model to navigate long-range dependencies, reconcile semantic conflicts, and perform iterative learning to arrive at a verified conclusion.

\subsection{Evaluation Methods}

Considering the diversity of tasks and their different evaluation focuses, we adopt the metrics categorized into three types: result-oriented, process-oriented, and open-ended. All scores are normalized to a range of 0 to 1. Metric details for each task are in Appendix~\ref{sec:capability}.

\textbf{Result-Oriented}\quad We evaluate the correctness of final outcomes via direct comparison using two metrics: \textit{accuracy} and \textit{precision}. Accuracy measures exact match with the complete set of correct answers, while precision measures the proportion of correct items in the model output, penalizing redundancy and errors \citep{precision}.

\textbf{Process-Oriented}\quad Process-oriented evaluation assesses the correctness and sequential consistency of procedural steps. We first use a stronger LLM to extract key steps, and then apply \textit{sequence order matching} based on the Longest Common Subsequence (LCS) to evaluate step coverage and ordering \citep{lcs}. Let the reference step sequence be $R = [r_1, r_2, \dots, r_m]$, and the extracted model-generated sequence be $\hat{A} = [\hat{a}_1, \hat{a}_2, \dots, \hat{a}_n]$. The sequence order matching score is defined as
\begin{equation}
S_{\mathrm{seq}} = \frac{\mathrm{LCS}(R, \hat{A})}{|R|}.
\end{equation}

\textbf{Open-Ended}\quad  
\textit{LLM-as-a-Judge} provides a flexible and scalable framework for evaluating open-ended subjective tasks~\citep{llm-as-a-judge,llmasajudge1}, effectively assessing semantic qualities such as consistency. Inspired by~\cite{weirocketeval}, each question in our work is associated with a set of evaluation criteria tailored to the task, ensuring that all relevant aspects of the answer are assessed.  
Given a question $Q$ and a model-generated answer $\hat{A}$, the final score is computed as
\begin{equation}
S_{llm} = \sum_{k=1}^{K} w_k \cdot J_{\theta}^{(k)}(Q, \hat{A}),
\end{equation}
where $J_{\theta}^{(k)}$ is the score for the $k$-th criterion and $w_k$ its relative weight.

{
\begin{table*}[!h]
\centering
\small

\begin{tabular}{c|ccccc|c|cc|c|c}
\hline

\multirow{2}{*}{\textbf{Model}} 
& \multicolumn{6}{c|}{\textbf{Declarative Memory}} 
& \multicolumn{3}{c|}{\textbf{Non-declarative Memory}} 
& \multirow{2}{*}{\textbf{Overall}} \\
\cline{2-7} \cline{8-10}

& \textbf{Ret.} 
& \textbf{Sum.} 
& \textbf{Iso.} 
& \textbf{Inf.} 
& \textbf{Rep.}
& \textbf{Avg}
& \textbf{Lea.} 
& \textbf{Hab.} 
& \textbf{Avg}
& \\
\hline

Gemini-3-Pro & 58.37 & 68.77 & 78.20 & \cellcolor{pink!30}{92.05} & 48.40 & 69.16 &47.37 &\cellcolor{pink!30}{84.16} & \cellcolor{pink!30}{65.77}  & \cellcolor{pink!30}{67.47}\\
GPT-5.1 & 76.41 & 68.77 & \cellcolor{pink!30}{82.87} & \cellcolor{green!10}72.38 & \cellcolor{pink!30}{63.60} & \cellcolor{pink!30}{72.81}  &\cellcolor{green!10}{50.86}  & 14.79 & 32.83  & \cellcolor{green!10}{52.82} \\
MiniMax-M2 & \cellcolor{green!10}{80.66} & \cellcolor{pink!30}{71.74} & \cellcolor{green!10}{82.27} & 59.38  & 61.06 & 71.02 &\cellcolor{pink!30}{51.35} &16.16 &\cellcolor{green!10}{33.76} & 52.39 \\
Mistral-Large & \cellcolor{pink!30}{81.11} & \cellcolor{green!10}{70.67} & 82.08 & 62.09& 61.26 & \cellcolor{green!10}{71.44} &48.10 & 18.42 & 33.26& 52.35 \\
DeepSeek-V3.2 & 72.68 & 67.00 & 76.91 & 60.84 & 57.26 & 66.94 &50.96 & 23.86 & 37.41  & 52.18\\
Kimi-K2 & 78.85 & 68.27 & 77.74 & 43.36 & \cellcolor{green!10}{62.26} &66.10 & 44.46 & 21.31 &32.89 & 49.50 \\
Llama-4-Maverick & 54.87 & 44.14 & 68.03 & 47.24  & 35.31 & 49.92& 25.01 & \cellcolor{green!10}{25.24} & 25.13 & 37.53 \\
\hline
Average & 71.85 & 65.52 &77.67 &62.48 &55.59 & 66.77 & 45.44 &29.13 & 37.29 & 52.03\\
Standard Deviation & 10.83 & 9.60 &5.01 &16.24 &10.30 & - & 9.34 &24.57 & - & - \\ 
\hline
\end{tabular}

\caption{Performance (\%) of evaluated LLMs. The abilities in Sec.~\ref{sec:memory_classification} are represented by their first three letters. {Avg} represents the average scores across corresponding sub-capabilities, and {Overall} refers to the averaged performance of both of them. Best scores in each column are highlighted in pink, and second-best scores in green. }
\label{tab:performance_llms}
\end{table*}
}

\section{Experiment Setup}

In this section, we elaborate on dataset construction details, and the LLMs and agent memory systems for evaluation. 

\subsection{Data Construction and Statistics}

To construct a diverse and representative benchmark, we selected data sources and appropriate LLMs for dialogue generation specified in Tab.~\ref{tab:source}. 

For topic-initiated generation, based on a survey of topic distributions in existing benchmarks~\citep{memoryagentbench,multichallenge,personamem,longmemeval} and the user interest report released by OpenAI~\citep{topic}, we identified 10 topic categories covering 50 specific topics, set LLM as DeepSeek-V3.2, $K$ as 5, $j$ as 5 in the planner stage, and in the generation phase, utilized 3 LLMs, which were respectively DeepSeek-V3.2~\citep{deepseek}, GPT-4.1~\citep{gpt4.1}, and Grok-4-Fast-Reasoning~\citep{grok4}. For narrative-inspired transformation, after assessing multiple datasets detailed in Appendix \ref{sec:text}, we selected Variable Tracking (VT) in Ruler as the source and employed LLM as DeepSeek-V3.2, $J$ as 9, and $T$ as 4.

In the end, EvolMem comprises 1,600 dialogues, with 6.82 sessions and 29.49 turns on average.

\subsection{LLMs For Evaluation}
We assess 7 LLMs with corresponding default settings for evaluation including DeepSeek-V3.2~\citep{deepseek}, GPT-5.1~\citep{gpt5}, Kimi-K2~\citep{kimik2}, Gemini-3-Pro~\citep{gemini3pro}, Mistral-Large~\citep{mistral_large}, MiniMax-M2~\citep{minimaxm2}, and Llama-4-Maverick~\citep{llama4}, which are all cutting-edge open-source and proprietary models.

\subsection{Agent Memory Systems For Evaluation}

Following prior taxonomies \citep{agent1,agent2}, memory agents can be grouped into structure-augmented RAG agents and Agentic memory agents based on their memory mechanisms.
For evaluation, we select representative agents from each class. Mem0 \citep{mem0} and HippoRAG \citep{hipporag} exemplify structure-augmented RAG agents, while MemoryOS \citep{memoryos} and A-MEM \citep{amem} represent agentic memory agents.

\section{Results and Analysis}

We present the evaluation results of LLMs and agent-based systems through specific evaluation methods and metrics shown in Appendix \ref{sec:evaluation} and Tab.~\ref{tab:source}. Accordingly, we analyze their performance, the impact of turns, the sensitivity of LLM generators, and the significance and evolutionary properties of our benchmark.

\subsection{Performance of LLMs}

The performance of LLMs are shown in Table~\ref{tab:performance_llms} with an illustration in Fig.~\ref{fig:llm_leaderboard}.

\textbf{Overall Performance}\quad The highest overall score is achieved by Gemini-3-Pro with 67.47\%, followed by GPT-5.1 and MiniMax-M2. Even the top-performing model exhibits notable weaknesses in certain dimensions, indicating that current LLMs lack a balanced and comprehensive memory ability across multiple tasks. For example, Gemini-3-Pro achieves 48.40\% in Reproduction and GPT-5.1 gets merely 14.79\% on Habituation.

\textbf{Capability Comparison}\quad As for fine-grained abilities, Isolation and Retrieval show stronger and more stable performance, with Isolation exhibiting the smallest inter-model variance of 5.01\%. In contrast, Habituation is the most challenging ability, yielding the lowest average performance and the largest variance. Notably, Gemini-3-Pro outperforms other models by approximately 4-5 times. This pronounced gap suggests that long-term context habituation ability has been largely overlooked in prior training and optimization.

\textbf{Model-Specific Analysis}\quad 
Gemini-3-Pro shows strong inference capabilities but underperforms in detailed retrieval tasks, as reflected by its relatively low retrieval score with 58.37\%. It also struggles with procedure-focused memorization tasks as reflected by Reproduction and Learning, indicating that its problem-solving relies heavily on internal parameters rather than leveraging dialogue history. In contrast, Mistral-Large and MiniMax-M2 exhibit the opposite pattern. They excel at recalling contextual content and cues, achieving higher scores in retrieval and reproduction. However, their inference abilities are comparatively weaker.

Overall, \textbf{the results indicate that LLMs perform better on declarative memory tasks than on non-declarative ones}, and \textbf{no model currently achieves all-round superiority}.
These findings also highlight the significance of EvolMem as a comprehensive benchmark. 

\begin{table*}[!h]
\centering
\small

\begin{tabular}{c|ccccc|c|cc|c|c}
\hline

\multirow{2}{*}{\textbf{Model}} 
& \multicolumn{6}{c|}{\textbf{Declarative Memory}} 
& \multicolumn{3}{c|}{\textbf{Non-declarative Memory}} 
& \multirow{2}{*}{\textbf{Overall}} \\
\cline{2-7} \cline{8-10}

& \textbf{Ret.} 
& \textbf{Sum.} 
& \textbf{Iso.} 
& \textbf{Inf.} 
& \textbf{Rep.}
& \textbf{Avg}
& \textbf{Lea.} 
& \textbf{Hab.} 
& \textbf{Avg}
& \\
\hline

Mem0 &
\cellcolor{green!10}{69.41} &
\cellcolor{green!10}{68.21} &
75.89 & 44.84 & 54.73 &
62.62 & 37.56 & 6.94 & 22.25 & 42.44 \\

HippoRAG &
42.54 & 43.68 & 61.20 &
\cellcolor{pink!30}{79.97} & 41.17 &
53.71 & 24.70 & 0.30 & 12.50 & 33.11 \\

MemoryOS &
63.95 &
\cellcolor{pink!30}{68.94} &
\cellcolor{pink!30}{77.19} &
\cellcolor{green!10}{64.05} &
\cellcolor{pink!30}{58.90} &
\cellcolor{green!10}{66.61} &
37.93 &
\cellcolor{green!10}{23.47} &
\cellcolor{green!10}{30.70} &
\cellcolor{green!10}{48.66} \\

A-MEM &
68.47 & 60.39 & 75.08 & 61.82 & 53.84 &
63.92 & 
\cellcolor{green!10}{40.93}& 9.37 & 25.15  & 44.54\\ 

\hline  

DeepSeek-V3.2 &
\cellcolor{pink!30}{72.68} & 67.00 &
\cellcolor{green!10}{76.91} &
60.84 &
\cellcolor{green!10}{57.26} &
\cellcolor{pink!30}{66.94} &
\cellcolor{pink!30}{50.96} &
\cellcolor{pink!30}{23.86} &
\cellcolor{pink!30}{37.41} &
\cellcolor{pink!30}{52.18} \\

\hline
\end{tabular}
\caption{Performance (\%) of memory agents with DeepSeek-V3.2 as the base model.}
\label{tab:memory_methods}
\end{table*}

\subsection{Performance of Agents}

We adopt DeepSeek-V3.2 as the shared base model of memory agents and evaluate their performance detailed in Appendix \ref{sec:agent} on EvolMem in Table~\ref{tab:memory_methods}.

\textbf{Effectiveness}\quad All evaluated \textbf{agent systems fail to surpass the base model}, DeepSeek-V3.2, which undermines the primary design objective of such memory modules. Among the agents, MemoryOS emerges as the best framework, followed by A-MEM and Mem0. When evaluated across diverse capability dimensions, agents display significant performance disparity. MemoryOS demonstrates superior capability among agents in tasks such as Summarization, Isolation, and Reproduction, outperforming the base model. Nevertheless, it displays significant weakness in Retrieval. Notably, while performing poorly across other capabilities, HippoRAG achieves the highest score in Inference. This performance discrepancy likely attributes to HippoRAG's graph-based PageRank mechanism, which effectively traces the relational chains essential for uncovering implicit logical dependencies.

The results reveal a substantial divergence in agent capability between the categories of declarative and non-declarative memory. While agents demonstrate comparable competence to the baseline in declarative tasks, their performance declines significantly in non-declarative memory. This difference demonstrates a critical deficiency of memory agents in procedural adaptation.

\textbf{Efficiency} We further analyze the efficiency of different memory mechanisms in Table~\ref{tab:memory_agents_speed}. The results show agentic memory systems incur substantial latency than structure-augmented RAG methods, with A-MEM requiring 16.3 times the duration of Mem0 system. This evidence suggests that \textbf{the iterative decision-making processes} inherent to autonomous memory evolution currently impose a prohibitive temporal cost, \textbf{significantly constraining their viability for real-time applications}.

\begin{table}[!t]
\centering

\begin{adjustbox}{width=\columnwidth}
\begin{tabular}{c|c c c c}
\hline
\textbf{Agents} & \textbf{mem0} & \textbf{HippoRAG} & \textbf{MemoryOS} & \textbf{A-MEM} \\
\hline
\textbf{Speed}$\downarrow$ & 30.17 & 92.76 & 183.24 & 492.60 \\
\hline
\end{tabular}
\end{adjustbox}
\caption{\textbf{Generation Speed} denotes average generation time (s) for each sample of different memory agents.}
\label{tab:memory_agents_speed}
\end{table}

\subsection{Feasibility of Using LLM Generators}
\begin{figure}[t]
  \centering
  \includegraphics[width=0.75\linewidth]{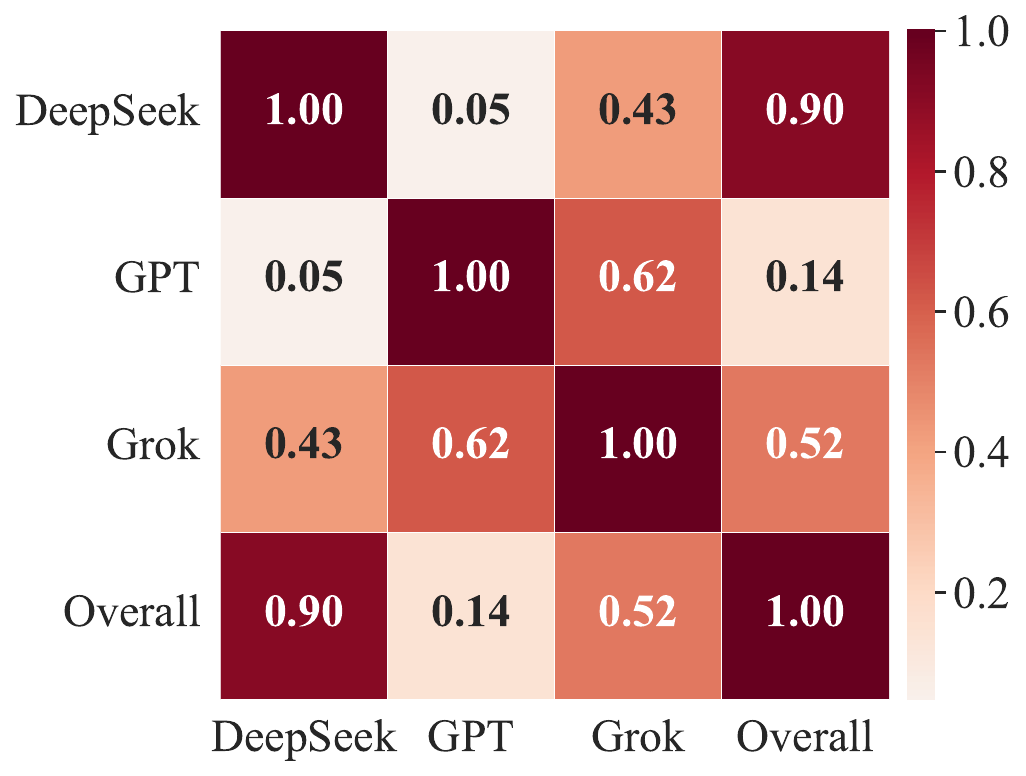}
  \caption{The Spearman's Correlation between the rankings under different generators and overall rankings.}
  \label{fig:generator}
\end{figure} 
Our proposed dialogue synthesis framework uses LLMs to generate data. To validate the feasibility of this approach, on the one hand, we conducted a manual verification process for a random sample of 50 instances from EvolMem. Three independent annotators assessed whether each Q\&A pair was grounded in and consistent with its dialogue context. The full agreement rate was 0.88, indicating consistent and reliable judgments. The results show that about 95.33\% of the synthesized data is valid.

On the other hand, we analyzed the sensitivity to different LLM generators on the results in Appendix~\ref{sec:generator}. We categorized the data in EvolMem according to its generator and compared model rankings across each subset. The results in Fig.~\ref{fig:generator} revealed that DeepSeek and Overall exhibited high consistency, with a Spearman correlation of 0.90, while GPT-4.1 showed a much lower correlation of 0.05. This discrepancy was due to differences in the topics retained after the complexity filter, as well as variations in model style and concentration. The Jaccard Similarity of topics between DeepSeek-V3.2 and GPT-4.1 is 83.53\%. These findings highlight the importance of using multiple LLM generators to ensure a diverse dataset, which is crucial for generating more accurate and fair model evaluations.

\subsection{Benchmark Evolution for Continual Assessment}
\begin{figure}[t]
  \centering
  \includegraphics[width=0.75\linewidth]{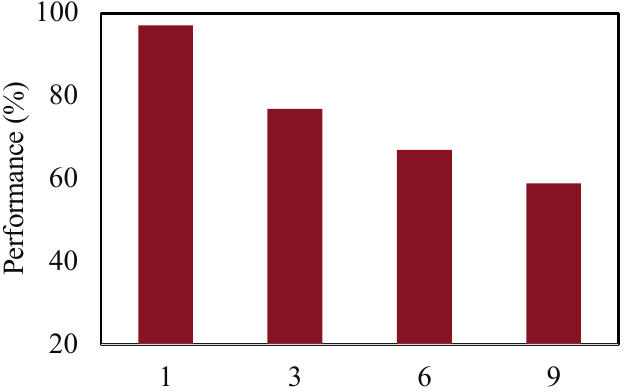}
  \caption{The performance of DeepSeek-V3.2 under different number of sessions.}
  \label{fig:evolution}
\end{figure}

The evolution of dataset difficulty is crucial for preventing performance saturation in long-term evaluations. To explore this, we conducted preliminary experiments by adjusting the complexity of individual sessions while constructing multi-session dialogues using topic-initiated generation. Experimental results, shown in Figure \ref{fig:evolution}, reveal that as the number of dialogue sessions increases, DeepSeek-V3.2's performance gradually declines. This suggests that EvolMem is capable of evolving in complexity, thereby maintaining its long-term effectiveness as model capabilities continue to advance.

\subsection{Correlation with Existing Benchmarks}
\begin{figure}[t]
  \centering
  \includegraphics[width=0.8\linewidth]{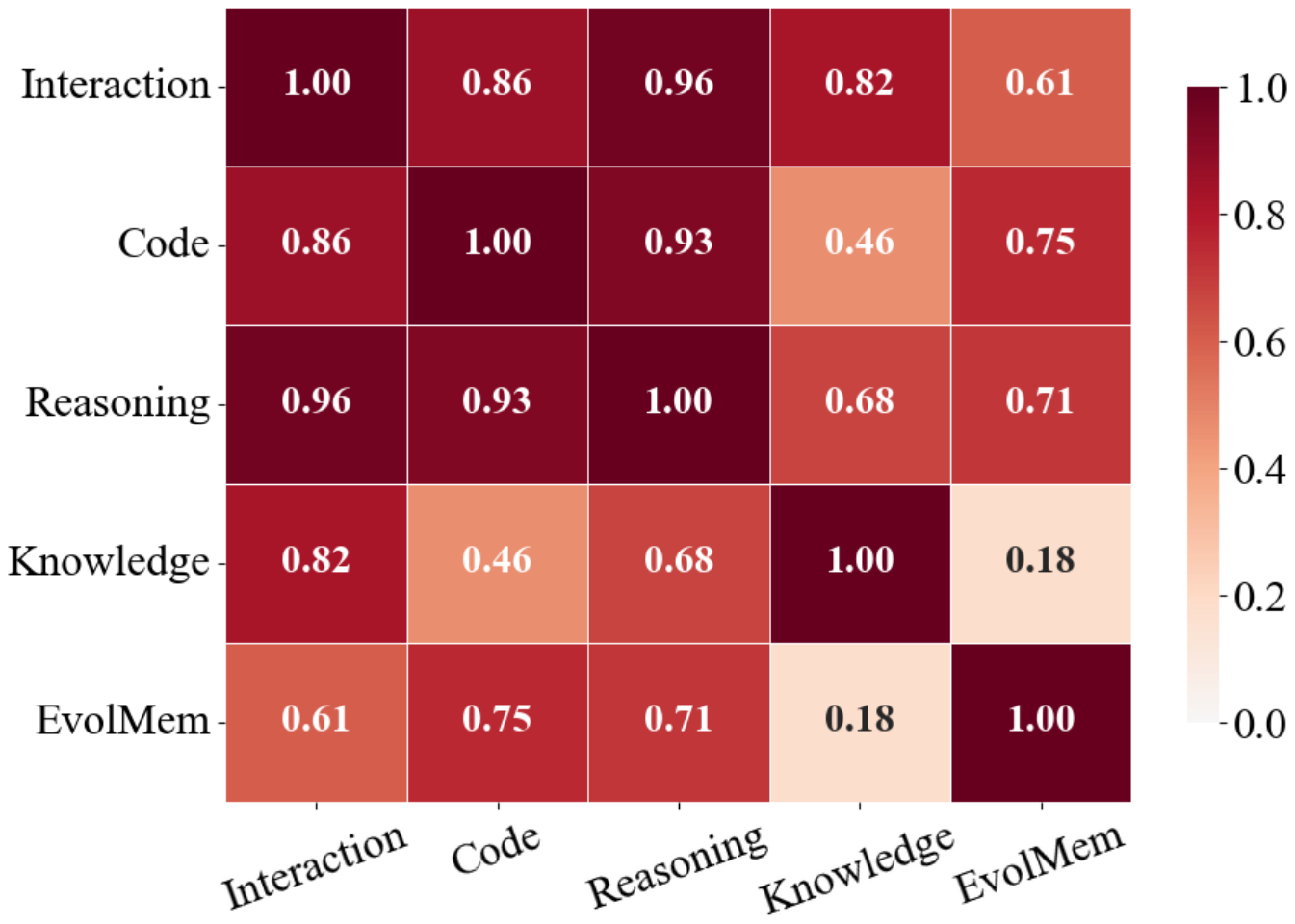}
   \caption{Spearman's Correlation between EvolMem and other four leaderboards.}
  \label{fig:leaderboard}
\end{figure}

Currently, there are many benchmarks and evaluation platforms~\citep{lmarena,AIBench} available to assess model capabilities. To distinguish memory performance from general capabilities, we integrate authoritative benchmarks—including SWE-Bench-Verified \citep{swe,llmstats2024}, AIME 2025 \citep{2025aime,llmstats2024}, and AGI-Eval \citep{agi-eval}, to evaluate LLM performance across code, reasoning, interaction, and knowledge domains. Spearman’s Correlation~\citep{srcc} between these rankings and the EvolMem ranking in Tab.~\ref{fig:leaderboard} shows that memory correlates most strongly with interaction at 0.61, suggesting memory is a critical component of interaction abilities. Strong correlations with Code at 0.75 and Reasoning at 0.71 indicate that improvements in these areas enhance memory performance. However, the low correlation with Knowledge at 0.18 reveals that EvolMem's memory capabilities are primarily driven by context understanding, rather than the out-of-context knowledge. This confirms that memory is an independent and essential dimension. 

\subsection{Case Study}

\begin{figure}[t]
  \centering
  \includegraphics[width=\linewidth]{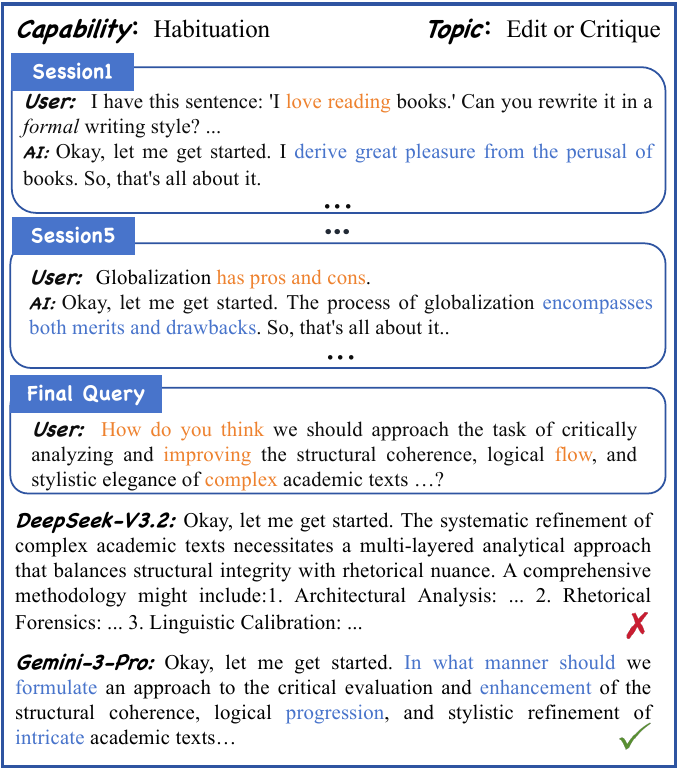}
  \caption{Case Study, where yellow and blue highlights indicate informal and formal wording.}
  \label{fig:casestudy}
\end{figure}

We select a sample from Habituation, which shows the largest performance variance, as a representative case study in Fig.~\ref{fig:casestudy}. In this example, the user consistently expects model to transfer style on the statements. Results exhibit that Gemini-3-Pro correctly preserves this long-term stylistic constraint and performs the expected style transfer perfectly. In contrast, DeepSeek-V3.2 ignores the habitual style preference encoded in the dialogue history and responds only to the current question, revealing a failure to capture persistent user intent in non-declarative memory.

\section{Conclusion}

We introduce EvolMem, a memory benchmark for both LLMs and agents. It is grounded in cognitive psychology and built using a multi-source, collaborative data synthesis framework to ensure diversity. The benchmark covers multi-faceted memory dimensions and uses various metrics for assessment. Our results show that models exhibit consistent weaknesses in non-declarative memory. Future work could explore strategies for improving agents' effectiveness and efficiency jointly, alongside multi-dimensional memory abilities.

\bibliography{main}
\newpage
\appendix
\appendix
\section{Topics for Dialogue Generation}
\label{sec:topics}
We utilize 50 topics from 8 different categories in Table.~\ref{tab:topics} for topic-initiated generation.
\begin{table}[!h]
  \centering
  \begin{tabular}{p{2.0cm}| p{4.5cm}}
    \hline
    \textbf{Category} & \textbf{Topics}  \\
    \hline
    \multirow{3}{*}{\shortstack{Technical\\ Help}}  & Computer Programming\\\cline{2-2}
    & Data Analysis \\ \cline{2-2}
    & Mathematical Calculation\\
    \hline

    \multirow{7}{*}{Writing} & Edit or Critique\\ \cline{2-2}
        & Fiction \\ \cline{2-2}
        & Argument \\ \cline{2-2}
        & Summary \\ \cline{2-2}
        & Survey \\ \cline{2-2}
        & Reports  \\ \cline{2-2}
        & Proposal \\ 
    \hline
    \multirow{5}{*}{\shortstack{Multidis-\\ciplinary\\ Studies}} & History \\ \cline{2-2}
    & Geography \\ \cline{2-2}
    & Law \\ \cline{2-2}
    & Physics \\ \cline{2-2}
    & Chemistry \\
    \hline
    \multirow{3}{*}{\shortstack{Professional\\ Development}} & Career Planning \\ \cline{2-2}
    & Time Management \\ \cline{2-2}
    & Self Management \\
    \hline
    \multirow{6}{*}{\shortstack{Communi-\\cation}} & Patient-Provider \\ \cline{2-2}
    & Healthcare Team \\ \cline{2-2}
    & Digital Health \\ \cline{2-2}
    & Email Drafts \\ \cline{2-2}
    & Memos \\ \cline{2-2}
    & Customer Service\\ 
    \hline
    \multirow{8}{*}{Economic} & Diet \\ \cline{2-2}
    & Trip \\ \cline{2-2}
    & Purchasable Products \\ \cline{2-2}
    & Cities \\ \cline{2-2}
    & Flight \\ \cline{2-2}
    & Stock \\ \cline{2-2}
    & Petrol \\ \cline{2-2}
    & Decoration \\ 
    \hline
    \multirow{7}{*}{Tutoring} & Biology Experiment \\ \cline{2-2}
    & Physics Experiment \\ \cline{2-2}
    & Chemistry Experiment \\ \cline{2-2}
    & Problem-Solving \\ \cline{2-2}
    & Common Sense \\ \cline{2-2}
    & Daily Skills \\ \cline{2-2}
    & Survival Skills \\ 
    \hline
    \multirow{3}{*}{Planning} & Meeting Schedules \\ \cline{2-2}
    & Book Outline \\ \cline{2-2}
    & Goout Schedules \\ 
    \hline
  \end{tabular}
  \caption{The specification of categories and topics utilized in prompt-driven dialogue generation.}
  \label{tab:topics}
\end{table}

\section{Implementation Details of Narrative-Inspired Transformation}
\label{sec:text}

\subsection{Task Mapping}

We select Ruler dataset due to its broad acceptance and rich collection of long-context instances. The dataset is organized into four primary categories: Needle in a Haystack (NIAH), Common Words (CWE), Question Answering (QA), and Variable Tracking (VT). For each category, we systematically analyze the challenges faced by evaluation models in memorizing long texts, identified the specific aspects of memorization being assessed, and reorganized the data according to our predefined task taxonomy.
\begin{table}
  \centering
  \begin{tabular}{p{1cm}p{6cm}}  
    \hline
    \textbf{Type} & \textbf{Characteristics} \\
    \hline
    NIAH  & Embedding key feature information in a large amount of lengthy text \\
    \hline
    CWE           & Providing a large number of words, some of which appear multiple times \\
    \hline
    QA     & Presenting rich information about people and events \\
    \hline
    VT      & Hiding variable assignments and changes in a large amount of irrelevant information \\
    \hline
  \end{tabular}
  \caption{Ruler dataset types and corresponding characteristics.}
  \label{tab:ruler}
\end{table}

As summarized in Table~\ref{tab:ruler}, each category targets a distinct core difficulty and corresponding capability. Specifically, NIAH evaluates retrieval ability, CWE focuses on summarization ability, QA assesses information isolation, and VT probes reasoning ability. 

\subsection{Session Transformation}
For each task, we sample 50 instances to conduct the evaluation on DeepSeek-V3.2. With the exception of VT, the model not only exhibits assessment capabilities consistent with those measured by existing dialogues from topic-initiated generation , but also achieves high performance, with scores exceeding 90\% on the remaining tasks.
 
We specifically explain \textbf{Challenge Injection}: 1) we find that incorporating contradictory information into VT is feasible and make it more difficult. Consequently, we deliberately insert a string inequality that conflicts with previous assignments, making several subsequent equations invalid. This type of design requires the model to backtrack and reason logically based on cross-turn historical information, identify and resolve inconsistencies in the context, and thus arrive at the correct conclusion. 2) we demonstrate the feasibility of incorporating ARC-style tasks, which represent a highly challenging form of implicit reasoning, into multi-turn dialogues. By deliberately embedding abstract patterns or spatial logic that demand more than mere surface-level recall, we force the model to perform probing inferences. This design necessitates that the model backtrack and integrate cross-turn historical information, identifying hidden logical structures within the context to arrive at the correct synthesis. 

\section{Details of EvolMem}
\label{sec:capability}
\subsection{Data Statistics}
\begin{figure}[t!]
  \centering
  \includegraphics[width=0.8\linewidth]{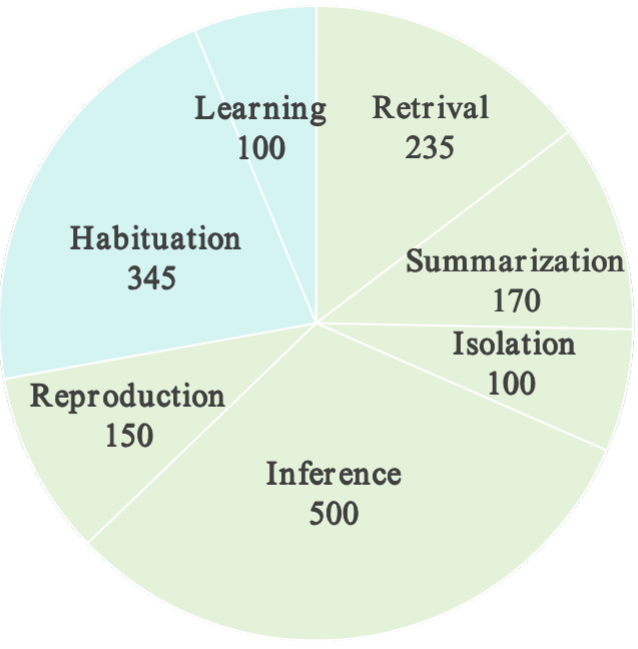}
  \caption{The distribution of capabilities in our benchmark. Green sectors represent declarative memory, while blue sectors represent non-declarative memory.}
  \label{fig:distribution}
\end{figure}

We display the distribution of samples for different capabilities of EvolMem in Fig.~\ref{fig:distribution}.

\subsection{Source \& Metric}
We construct our multi-session dialogues through topic-initiated generation (TIG) and narrative-inspired transformation (NIT), the source and evaluation metrics of capabilities are shown in Tab.~\ref{tab:source}.

\begin{table}[t!]
  \centering
  \begin{tabular}{p{2cm}p{1.8cm}p{2.5cm}}  
    \hline
    \textbf{Capability} & \textbf{Source} & \textbf{Metric}\\
    \hline
    Retrieval  & TIG & Acc\& LLM \\
    \hline
    Summarization  & TIG & Acc \& LLM \\
    \hline
    Isolation     & TIG & Acc \\
    \hline
    Inference      & NIT & Prec\\
    \hline
    Reproduction & TIG & LCS \& LLM \\
    \hline
    Learning & TIG \& NIT & Acc \& LCS \\
    \hline
    Habituation & TIG & Acc \& LLM \\
    \hline
  \end{tabular}
  \caption{The corresponding source and evaluation metrics of each capability. Acc represents Accuracy, Prec represents Precision, and LLM refers to the LLM judge. }
  \label{tab:source}
\end{table}

\section{Memory Agents}
\label{sec:agent}
This section details the memory mechanisms and experimental implementation of the memory agents considered in this work.

\subsection{Memory Mechanisms}
Memory agents are LLM-based agents equipped with explicit memory mechanisms that enable information to be retained and leveraged beyond the fixed context window, thus alleviating the inherent context-length constraint during task execution. 

\textbf{Mem0}
introduces a modular two-phase memory architecture that incrementally extracts and maintains salient information from dialogues. In the extraction phase, an LLM proposes candidate memory items by analyzing the current interaction alongside a global conversation summary, which is asynchronously generated by an LLM to capture the evolving context. These candidates are then matched against existing entries via vector retrieval; subsequently, an LLM-driven tool-calling interface determines whether to insert, merge, delete, or discard each item, ensuring a compact and consistent memory store. A variant augmented with graphs further encodes memories as entity–relation triplets, enabling retrieval over richer relational structures than flat textual chunks.

\textbf{HippoRAG}
realizes structure-augmented RAG by converting the corpus into an open knowledge graph via LLM-based OpenIE, then performing Personalized PageRank over this graph to support associative, multi-hop retrieval. At query time, named entities are assigned to graph nodes that serve as seeds in the PageRank process, propagating probability mass through relation edges to identify passages that occur along high-scoring paths. This dense–sparse integration yields a retrieval signal that captures both local semantic similarity and longer relational chains, improving factual, sense-making and associative memory tasks over standard embedding-based RAG.

\textbf{A-MEM}
 introduces an agentic memory architecture in which interaction traces are encoded as LLM-generated ``notes'' comprising content, timestamps, keywords, tags, contextual descriptions and embeddings. A new note starts a two-phase agentic routine: first, similarity-based retrieval identifies candidate neighboring notes; second, an LLM controller performs link construction and memory evolution, revising existing notes and updating their attributes in light of new information. Drawing on Zettelkasten-style personal knowledge management, A-Mem builds a graph of interconnected notes that yields a self-evolving memory structure, supporting both fine-grained semantic retrieval and higher-level knowledge organization for LLM agents.

 \textbf{MemoryOS}
 models an agent’s memory as an OS-like hierarchical framework, coordinating multi-tier storage (short-/mid-/long-term). Dialogue input stream is first held in a short-term buffer, periodically condensed into topic-level units in mid-term memory, and subsequently promoted to long-term memory based on FIFO and ``heat''-style signals that reflect recency and usage. Specialized retrieval and generation components perform semantic search over these tiers and re-inject selected content into the LLM’s context, supporting stable persona tracking and coherent behavior over prolonged interactions.

 \subsection{Implementation Details}
 In this section, we detail the experimental setup of different memory agents. To ensure a fair and robust benchmark, we standardize the backbone model, embedding model, and key hyperparameters across all agents. We utilize DeepSeek-V3.2 as the Backbone LLM and all-MiniLM-L6-v2 as the embedding model for all agents.

For Mem0, we utilize a memory search limit of $s=10$, and sliding window of the last $m=10$ messages alongside the retrieved memories, which aligns with the original paper's settings.

For HippoRAG, the Synonym Threshold is set to 0.8, and we use $top$-$k=10$ seed nodes for initial linking, 200 passage nodes via the Personalized PageRank (PPR) algorithm and $top$-$k=10$ passages as the context for final generation.

For MemoryOS, we use a fixed Short-Term Memory (STM) queue size of 7 and a Long-Term Personal Memory (LPM) capacity of 100 entries. The Heat threshold is set to $\tau=5$, the semantic similarity threshold $\theta=0.6$, and for retrieval, we use $top$-$k=10$ queues. All parameters strictly follow the original experimental setup.

For A-MEM, we use the default insertion and query settings, selecting $top$-$k=10$ semantically similar notes and $top$-$k=10$ relevant notes for retrieval respectively.

\section{Evaluation Plan}
\label{sec:evaluation}
We outline a unified evaluation protocol that maps each fine-grained memory ability to a tailored scoring strategy, balancing objective matching with judge-based assessment when needed.

For Retrieval, Summarization, and Isolation, we utilize LLM-as-a-Judge, and directly let the judge model output the scoring results according to the corresponding evaluation pipeline.

For Habituation, where correctness is defined by whether the model follows the initial instructions rather than matching a fixed answer, we adopt a hybrid of LLM-as-a-Judge and threshold approach: the judge model first produces continuous scores via the evaluation pipeline, then binarizes them based on a preset threshold 0.8. The evaluation metric for this task is Instruction Compliance, reflecting whether the model correctly follows the given instructions. 

For Reproduction, although the evaluation pipeline has a certain degree of reliability in terms of content consistency, it cannot fully determine whether the step order is correct. Therefore, we adopt a structured sequence order matching, including step extraction, reverse order detection, comparing the similarity between the model's step sequence and the correct step sequence and calculating the longest common subsequence (LCS) ratio.

For Inference, we utilize direct comparison. Since the model's answers contain redundant expressions, we first require a strong model to extract answers and normalize the format to make it consistent with the correct answer. Then, we directly compare the normalized content to obtain the final score.

For Learning, we utilize respectively direct comparison and Sequence order matching. For ARC-style tasks, we normalize model outputs and compute accuracy by exact match against the ground-truth answer. For topic-initiated dialogues, we first use an LLM to extract the procedural sequence, compute TF-IDF similarity, and then calculate the LCS score.

\textbf{LLM-as-a-Judge Feasibility} involves manual verification of the correctiveness of using LLM-as-a-Judge. We randomly select 50 evaluation samples from the dataset and distribute them to experts for verification, and the feasibility rate is over 92.67\%.

\section{Sensitivity of LLM Generators}
\label{sec:generator}
The specific score of test LLMs under different generation LLMs and the rank compared with EvolMem are shown in Table \ref{tab:generator}. These results highlight the importance of using multiple LLM generators to ensure a diverse dataset, which is crucial for generating more accurate and fair model evaluations.

\definecolor{darkgreen}{rgb}{0.0,0.5,0.0}
\begin{table}[!h]
  \centering
  \small
  \begin{tabular}{m{2.4cm} | m{1.0cm} m{1.0cm} m{1.0cm}}
    \hline
   \textbf{Models} & \textbf{DeepSeek} &\textbf{GPT}  & \textbf{Grok} \\
    \hline
    Gemini-3-Pro & 67.81\% & 61.93\%\textcolor{red}{$\downarrow$3} & 65.71\%\textcolor{red}{$\downarrow$2} \\
    GPT-5.1 & 67.29\% & 60.69\%\textcolor{red}{$\downarrow$3} & 65.41\%\textcolor{red}{$\downarrow$2}\\
    MiniMax-M2 & 65.55\%\textcolor{red}{$\downarrow$1} & 62.40\%\textcolor{darkgreen}{$\uparrow$1}  & 67.15\%\textcolor{darkgreen}{$\uparrow$2} \\
    Mistral-Large & 66.81\%\textcolor{darkgreen}{$\uparrow$1} & 62.18\%\textcolor{darkgreen}{$\uparrow$1}  & 66.51\%\textcolor{darkgreen}{$\uparrow$2}\\
    DeepSeek-V3.2 & 63.49\% & 57.76\%\textcolor{red}{$\downarrow$1} &60.76\%\textcolor{red}{$\downarrow$1} \\
    Kimi-K2 & 63.38\% & 62.79\%\textcolor{darkgreen}{$\uparrow$5} & 63.49\%\textcolor{darkgreen}{$\uparrow$1}  \\
    Llama-4-Maverick &46.84\% & 42.30\% & 45.65\% \\
    \hline
   \textbf{SRCC} &96.43\% & 17.86\% & 67.86\% \\
    \hline
  \end{tabular}
  \caption{Memory comprehensive performance of models in different dialogue generators and the change of rankings compared with overall score in Tab.2. \textbf{DeepSeek} represents DeepSeek-V3.2. \textbf{GPT} represents GPT-4.1. \textbf{Grok} represents Grok-4-Fast-Reasoning.}
  \label{tab:generator}
\end{table}

\end{document}